\documentclass[11pt,a4paper]{article}
\usepackage[hyperref]{emnlp-ijcnlp-2019}
\usepackage{times}
\usepackage{latexsym}
\usepackage{url}
\usepackage{tikz}
\usetikzlibrary{positioning}
\usetikzlibrary{decorations.pathreplacing}
\usetikzlibrary{calc}
\usepackage{graphicx}  %Required
\usepackage{float}
\usepackage{tabu,multirow}
\usepackage{url}

\aclfinalcopy % Uncomment this line for the final submission

\title{AMPERSAND: Argument Mining for PERSuAsive oNline Discussions}

\author{Tuhin Chakrabarty\textsuperscript{1}, 
  Christopher Hidey\textsuperscript{1}, 
  Smaranda Muresan\textsuperscript{1,2}\\,
  \textbf{Kathleen Mckeown}\textsuperscript{1} \textbf{and}
  \textbf{Alyssa Hwang}\textsuperscript{1} \\
  \textsuperscript{1}Department of Computer Science, Columbia University \\
  \textsuperscript{2}Data Science Institute, Columbia University\\\AND
  {\tt \{tuhin.chakrabarty, smara, a.hwang\}@columbia.edu} \\
  {\tt \{chidey, kathy\}@cs.columbia.edu} 
  }
\date{}

\begin{document}
\maketitle
\begin{abstract}

Argumentation is a type of discourse where speakers try to persuade their audience about the reasonableness of a claim by presenting supportive arguments. Most work in argument mining has focused on modeling arguments in monologues. We propose a computational model for argument mining in online persuasive discussion forums that brings together the micro-level (argument as product) and macro-level (argument as process) models of argumentation. 
Fundamentally, this approach relies on identifying relations between components of arguments in a discussion thread.  Our approach for relation prediction uses contextual information in terms of fine-tuning a pre-trained language model and leveraging discourse relations based on Rhetorical Structure Theory.  We additionally propose a candidate selection method to automatically predict what parts of one's argument will be targeted by other participants in the discussion. Our models obtain significant improvements compared to recent state-of-the-art approaches using pointer networks and a pre-trained language model.
  
\end{abstract}

\section{Introduction}

Argument 
mining
is a field of corpus-based discourse analysis that involves the automatic identification of argumentative structures in text. Most current studies have focused on 
monologues
or \textit{micro-level} models of argument that aim to identify the structure of a single argument by identifying the argument components (classes such as ``claim'' and ``premise'') and relations between them (``support'' or ``attack'')  \citep{somasundaran2007detecting,stab2014identifying,swanson2015argument,feng2011,habernalgurevych2017,peldszusstede2015}. \textit{Macro-level} models (or dialogical models) and rhetorical models which focus on the \emph{process} of argument in a dialogue \citep{bentaharetal2010} have received less attention.  

\begin{figure}
\includegraphics[scale=0.4, trim={0 1cm 0 0},]{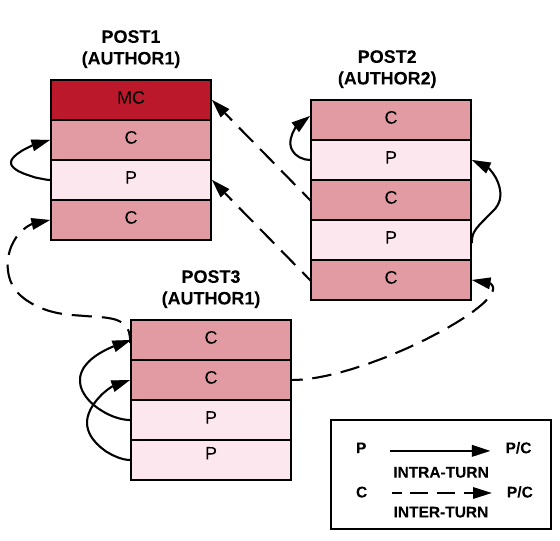}
\caption{\label{thread_scheme} Annotated Discussion Thread (C: Claim, P: Premise, MC: Main Claim)}
\end{figure}
We propose a novel approach to automatically identify the argument structure in persuasive dialogues that brings together the micro-level and the macro-level models of argumentation. Our approach identifies argument components in a full discussion thread and two kinds of argument relations: \emph{inter-turn relations} (argumentative relations to support or attack the other person's argument) and \emph{intra-turn} relations (to support one's claim). 
Figure \ref{thread_scheme} shows a thread structure consisting of multiple posts with argumentative components (main claims, claims or premises) and both intra- and inter-turn relations.
We focus here on the relation identification task (i.e, predicting the existence of a relation between two argument components). 

To learn to predict relations, we \textbf{annotate argumentative relations} in threads from the Change My View (CMV) subreddit. In addition to the CMV dataset we also \textbf{introduce two novel distantly-labeled data-sets} that incorporate the macro- and micro-level context (further described in Section \ref{data}). We also \textbf{propose a new transfer learning approach} to fine-tune a pre-trained language model \citep{bert} on the distant-labeled data (Section \ref{methods}) and demonstrate improvement on both argument component classification and relation prediction (Section \ref{results}).   We further show that \textbf{using discourse relations} based on Rhetorical Structure Theory \citep{mannRST} improves the results for relation identification both for inter-turn and intra-turn relations. Overall, our approach for argument mining \textbf{obtains statistically significant improvement} over a state-of-the-art model based on pointer networks 
\citep{japanese}
and a strong baseline using a pre-trained language model \citep{bert}.
We make our data,\footnote{https://github.com/chridey/change-my-view-modes} code, and trained models publicly available.\footnote{https://github.com/tuhinjubcse/AMPERSAND-EMNLP2019}

\section{Related Work}
 
\paragraph{Argument mining on monologues}
Most prior work in argument mining (AM) has focused on monologues or essays. \newcite{pe} and \newcite{essay} used pipeline approaches for AM, combining parts of the pipeline using integer linear programming (ILP) to classify argumentative components. \newcite{pe} represented relations of argument components in essays as tree structures. Both \newcite{pe} and \newcite{essay} propose joint inference models using ILP to detect argumentative relations. They however rely on handcrafted (structural, lexical, indicator, and discourse) features. We on the other hand use a transfer learning approach for argument mining in discussion forums.

Recent work has examined neural models for argument mining. \newcite{potash} use pointer networks \citep{pointer2015} to predict both argumentative components and relations in a joint model.
For our work, we focus primarily on relation prediction but conduct experiments with predicting argumentative components in a pipeline.
A full end-to-end neural sequence tagging model was developed by \newcite{end2end} that predicts argumentative discourse units (ADUs), components, and relations at the token level. In contrast, we assume we have ADUs and predict components and relations at that level.

\paragraph{Argument mining on discussion forums (online discussion)}

Most computational work related to argumentation as a process has focused on the detection of agreement and disagreement in online interactions \citep{abbott2011can, sridhar2015joint, rosenthal2015couldn, walker2012corpus}. However, these approaches do not identify the argument components that the (dis)agreement has scope over (i.e., what has been targeted by a disagreement or agreement move).
Also in these approaches, researchers predict the type of relation (e.g. agreement) given that a relationship exists. On the contrary, we predict both argument components as well as the existence of a relation between them. \newcite{stancebackup} and \newcite{murakami} address relations between complete arguments but without the micro-structure of arguments as in \newcite{pe}. \newcite{target} introduce a scheme to annotate inter-turn relations between two posts as ``target-callout'', and intra-turn relations as ``stance-rationale". However, their empirical study is reduced to predicting the type of inter-turn relations as agree/disagree/other. Our computational model on the other hand handles both macro- and micro- level structures of arguments (argument components and relations). 

\newcite{budzynska2014towards} focused on building foundations for extracting argument structure from dialogical exchanges (radio-debates) in which that structure may be implicit in the dynamics of the dialogue itself. By combining recent advances in theoretical understanding of inferential anchors in dialogue with grammatical techniques for automatic recognition of pragmatic features, they produced results for illocutionary structure parsing which are comparable with existing techniques acting at a similar level such as rhetorical structure parsing. Furthermore,  \newcite{Visser2019} presented US2016, the largest publicly available set of corpora of annotated dialogical argumentation. Their annotation covers argumentative relations, dialogue acts and pragmatic features.

Although \newcite{factor-graph} tried relation prediction between arguments in user comments on web discussion forums using structured SVM and RNN the  work closest to our task is that of \newcite{japanese}. They propose a pointer network model that takes into consideration the constraints of thread structure. Their model discriminates between types of arguments (e.g., claims or premises) and both intra-turn and inter-turn relations, simultaneously. Their dataset, which is not publically available, is three times larger than ours, so instead we focus on transfer learning approaches that take advantage of discourse and dialogue context and use their model as our baseline.

\section{Data}\label{data}

\subsection{Labeled Persuasive Forum Data}
\label{labeled_data}

To learn to predict relations, we use a corpus of 78 threads from the CMV subreddit annotated with argumentative components~\citep{cmv}. 
%KM-final - I have fixed the one above to be correct. Take a look at how I cite it and how it appears in the paper. 
The authors annotate \emph{claims} (``propositions that express the speaker’s stance on a certain matter''), 
and \emph{premises} (``propositions that express a justification provided by the speaker in support of a claim"). In this data, the \emph{main claim}, or the central position of the original poster, is always the title of the original post.

We expand this corpus by annotating the argumentative relations among these propositions (both inter-turn and intra-turn) and 
extend the corpus by annotating additional argument components (using the guidelines of the authors) for a total of 112 threads. It is to be noted that the  claims, premises, and relations have been annotated jointly. Our annotation results on the argumentative components for the additional threads were similar to \newcite{cmv} ( IAA using Kripendorff alpha is $0.63$ for claims and $0.65$ for premises). The annotation guidelines for relations are available in the aforementioned Github repository.

\paragraph{Intra-turn Argumentative Relations} As in previous work \citep{japanese}, 
%KM-final - fixed above citation.
we restrict intra-turn relations to be between a premise and another claim or premise, where the premise either supports or attacks the claim or other premise. Evidence in the form of a premise is either  support or  attack. Consider the following example:

\begin{quote}
[Living and studying overseas is an irreplaceable experience.]\textsubscript{0:CLAIM} [\color{red}One will struggle with loneliness\color{black}]\textsubscript{1:PREMISE:ATTACK:0} [\color{red}but those difficulties will turn into valuable experiences.\color{black}]\textsubscript{2:PREMISE:ATTACK:1} [\color{blue}Moreover, one will learn to live without depending on anyone.\color{black}]\textsubscript{3:PREMISE:SUPPORT:0}
\end{quote}

The example illustrates that the argumentative component at index 0 is a claim, and is followed by attacking one's own claim, and in turn attacking that premise (an example of the rhetorical move known as prolepsis or prebuttal). 
They conclude the argument by providing supporting evidence for their initial claim.

\paragraph{Inter-turn Argumentative Relations} Inter-turn relations
connect the arguments of two participants in the discussion (agreement or disagreement). The argumentative components involved in inter-turn relations are claims, 
as the nature of dialogic argumentation is a difference in stance. An \emph{agreement} relation expresses an agreement or positive evaluation of another user's claim whereas a \emph{disagreement/attack} relation expresses disagreement.
We further divide disagreement/attack relations into \emph{rebuttal} and \emph{undercutter} types to distinguish when a claim directly challenges the truth of the other claim or challenges the reasoning connecting the other claim to the premise that supports it, respectively. We also allowed annotators to label claims as \emph{partial} agreement or disagreement/attack if the response concedes part of the claim, depending on the order of the concession.

In the following example, User A makes a claim and supports it with a premise. User B agrees with the claim but User C disputes the idea that there even is global stability. Meanwhile, User D disagrees with the reasoning that the past is always a good predictor of current events.
\begin{quote}
A: [I think the biggest threat to global stability comes from the political fringes.]\textsubscript{0:CLAIM} [It has been like that in the past.]\textsubscript{1:PREMISE}

B: [\color{blue}Good arguments.\color{black}]\textsubscript{2:AGREEMENT:0}

C: [\color{red}The only constant is change.\color{black}]\textsubscript{3:REBUTTAL:0}

D: [\color{olive}What happened in the past has nothing to do with the present.\color{black}]\textsubscript{4:UNDERCUTTER:1}
\end{quote}

We obtain moderate agreement for relation annotations, similar to other argumentative tasks \citep{japanese}. The Inter-Annotator Agreement (IAA) with Kripendorff's $\alpha$ is 0.61 for relation presence and 0.63 for relation types.

\begin{table*}[]
\begin{center}
\small{
\begin{tabu}{|l|l|}
\hline
\multicolumn{2}{|l|}{\begin{tabular}[c]{@{}l@{}}\textbf{CMV: A rise in female representation in elected government isn't a good or bad thing}.\\ According to this new story, a record number of women are seeking office in this year's US midterm elections.\\ While some observers hail this phenomenon as a step in the right direction, \color{red}{I don't think it's good thing one way or the }\\ \color{red}{other: a politician's sex has zero bearing on their ability to govern or craft effective legislation.} \color{black}As such... 
\end{tabular}}                                                                                 \\ \hline
\multicolumn{2}{|l|}{\begin{tabular}[c]{@{}l@{}}\color{blue}``I don't think it's good thing one way or the other: a politician's sex has zero bearing on their ability to govern or craft \\\color{blue}effective legislation'' \color{black}Nobody is saying that  women are better politicians than men, and thus, more female representation\\ is inherently better for our political system. Rather, the argument is that... 
\end{tabular}} \\ \hline
\end{tabu}
}
\end{center}
\caption{\label{summdata} Two posts from the CMV sub-reddit where a claim is targeted by the response}
\end{table*}

In total, the dataset contains 380 turns of dialogue for 2756 sentences.
There were 2741 argumentative propositions out of which 1205 are claims and 1536 are premises, with an additional 799 non-argumentative propositions.
66\% of relations were in support, 26\% attacking, and 8\% partial.
As we found that most intra-turn relations were in support and inter-turn relations were attacking, due to the dialogic nature of the data, for our experiments we only predicted whether a relation was present and not the relation type.

Overall, there are 7.06 sentences per post for our dataset, compared to 4.19 for \newcite{japanese}.  This results in a large number of possible relations, as all pairs of argumentative components are candidates.  The resulting dataset is very unbalanced (only 4.6\% of 27254 possible pairs have a relation in the intra-turn case with only 3.2\% of 26695 for inter-turn), adding an extra challenge to modeling this task.

\subsection{Distant-Labeled Data}
\label{distant_data}

As the size of this dataset is small for modern deep learning approaches, we leverage distant-labeled data from Reddit and use transfer learning techniques to fine-tune a model on the appropriate context --- micro-level for intra-turn relations and macro-level (dialogue) for inter-turn relations.

\paragraph{Micro-level Context Data}

In order to leverage transfer learning methods, we need a large dataset with distant-labeled relation pairs. In previous work, \newcite{imho} collected a distant-labeled corpus of opinionated claims using sentences containing the internet acronyms IMO (in my opinion) or IMHO (in my humble opinion) from  Reddit. We leverage their data to model relation pairs by considering the following sentence (i.e., a premise) as well (when it was present), resulting in a total of 4.6 million comments. We denote this dataset as \textbf{IMHO+context}. The following example shows an opinionated claim (in bold) backed by a supporting premise (in italics).

\begin{quote}

\textbf{IMHO, Calorie-counting is a crock what you have to look at is how wholesome are the foods you are eating.} \emph{Refined sugar is worse than just empty calories - I believe your body uses a lot of nutrients up just processing and digesting it.}
\end{quote}
We assume in this data that a relation exists between the sentence containing IMHO and the following one.  While a premise does not always directly follow a claim, it does so frequently enough that this noisy data should be helpful.

\paragraph{Macro-level Context Data}
While the IMHO data is useful for modeling context from consecutive sentences from the same author, inter-turn relations are of a dialogic nature and would benefit from models that consider that macro-level context. We take advantage of a feature of Reddit: when responding to a post, users	can easily \textit{quote} another user's response, which shows up in the metadata. Particularly in CMV, this feature is used to highlight exactly what part of someone's argument a particular user is targeting. In the example in Table \ref{summdata}, the response contains an exact quote of a claim in the original post. We collect 95,406 threads from the full 
CMV subreddit between 2013-02-16 and 2018-09-05 and find pairs of posts where the quoted text	in the response is an exact match for the original post (removing threads that overlap with the labeled data). This phenomenon occurs a minority of the time, but we obtain 19,413 threads. When the quote feature is used, posters often respond to multiple points in the original text, so for the 19,413 threads we obtain 97,636 pairs. As most language model fine-tuning is performed at the sentence level, we take the quoted text and the  following sentence as our distant-labeled inter-post pairs. We refer to this dataset as \textbf{QR}, for quote-response pairs.

\begin{figure}[t]
\begin{center}
\includegraphics[scale=0.55]{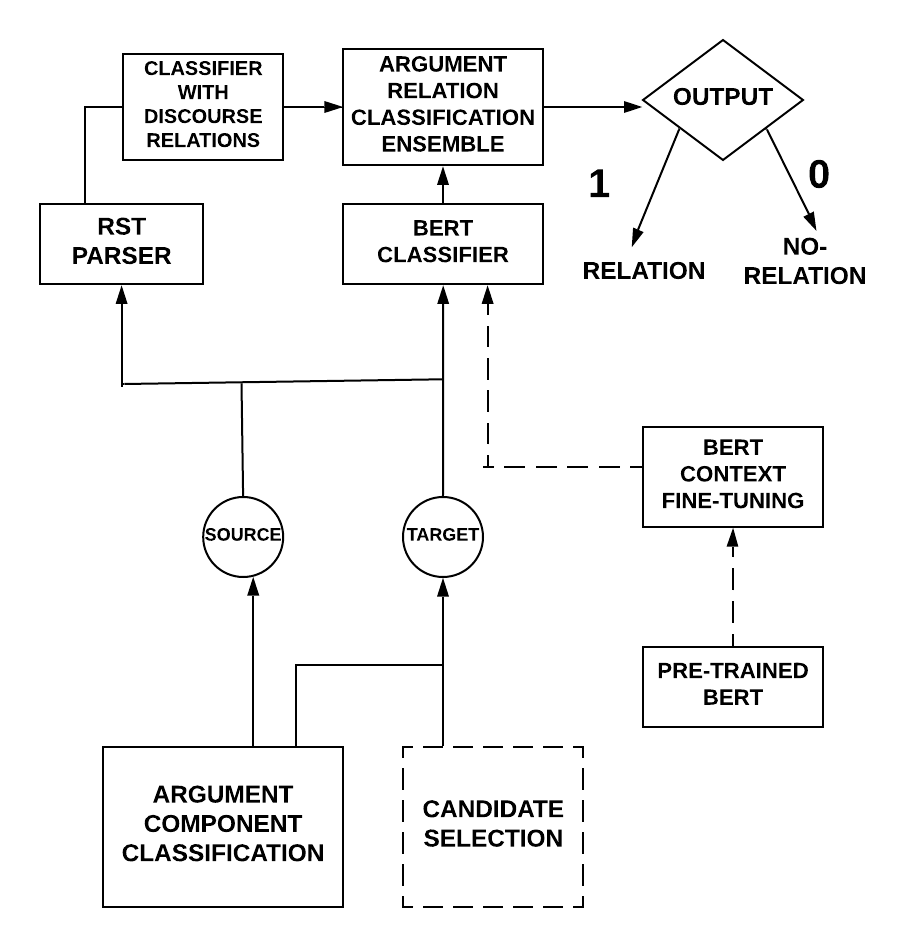}
\label{fig:fig1}
\caption{\label{modelpic} Schematic of our pipeline involving various components. The Candidate Selection Component is only used for Inter-turn relation identification}
\end{center}
\end{figure}

\section{Methods} \label{methods}

Identifying argumentative components is a necessary precursor to predicting an argumentative relation.  In our data, we require the ``source'' of an intra-turn relation to be a premise that supports or attacks a ``target'' (a premise or another claim).  For inter-turn relations, the source is always a claim that agrees or disagrees with a target.
Thus, we model this process as a pipeline: perform three-way classification on claims, premises, and non-arguments and then predict if an outgoing relation exists from the source premise/claim to a target premise/claim. In predicting these relations, we consider all possible source-target pairs of premises and argumentative components within a single post (for intra-turn) and claims from one post and argumentative components from another post (for inter-turn).

As our data set is relatively small, we leverage recent advances in transfer learning for natural language processing. BERT \citep{bert} has shown excellent performance on many semantic tasks, both for single sentences and pairs of sentences, making this model an ideal fit for both argument component classification and relation prediction. 
The BERT model is initially trained with a multi-task objective (masked language modeling and next-sentence prediction) over a 3.3 billion word English corpus. In the standard use, given a pre-trained BERT model, the model can be used for transfer learning by fine-tuning on a domain-specific corpus using a supervised learning objective.

We take an additional step of fine-tuning on the relevant distant-labeled data from Section \ref{distant_data} before again fine-tuning on our data from Section \ref{labeled_data}.  We hypothesize that this novel use of BERT will help because the data is structured such that the sentence and next sentence pair (for intra-turn) or quote and response pair (for inter-turn) will encourage the model to learn features that improve performance on detecting argumentative components and relations.
For argumentative components, we use this fine-tuned BERT model to classify a single proposition (as BERT can be used for both single inputs and input pairs).  For relations (both intra-turn and inter-turn), we predict the relation between a pair of propositions. In the relation case, we postulate that additional components besides BERT can help: adding features derived from RST structure and pruning the space of possible pairs by selecting candidate target components. The pipeline is illustrated in Figure \ref{modelpic}.

\paragraph{Argument Component Classification}
We fine-tune a BERT model on the \textbf{IMHO+context} dataset described in Section \ref{distant_data} using both the masked language modeling and next sentence prediction objectives. We then again fine-tune BERT to predict the argument components for both sources and targets, as indicated in Figure \ref{modelpic}.
Given the predicted argumentative components, we consider all possible valid pairs either within a post (for intra-turn) or across (inter-turn) and make a binary prediction of whether a relation is present.

\paragraph{Context Fine-Tuning} For intra-turn relation prediction we use the same fine-tuned BERT model on \textbf{IMHO+context} that we used for argument component classification, as premises often immediately follow claims so this task is a noisy analogue to the task of interest.  
%KM2 - what dataset do you use to do this? And why do you have to do this second fine-tuning? And is it all possible pairs of sentences in the corpus where some have relations and some do not?
We then fine-tune on the relation prediction task on all possible pairs, using the labeled relations in the CMV data. 
For inter-turn relation prediction, %rather than use \textbf{IMHO+context}, which consists of consecutive sentences from the same author, 
we fine-tune first on the \textbf{QR} dataset, where the dialogue context more closely represents our labeled inter-post relations. Then, we fine-tune on inter-turn relation prediction using all possible pairs as training. This process is indicated in Figure \ref{modelpic}, where we use the appropriate context fine-tuning to obtain a relation classifier.

\paragraph{Discourse Relations}

% \begin{table} [h!]
% \small
% \centering
% \begin{tabular}{|c|c|}
% \hline
% ARG1    & ARG2                \\ \hline
% \begin{tabular}[c]{@{}l@{}}(If existence from your \\perspective)( lies solely\\ on your consciousness)\end{tabular} & \begin{tabular}[c]{@{}l@{}}(after you die)( it doesn't\\ matter what you left)\end{tabular} \\ \hline
% \end{tabular}
% \caption{Example of an argumentative relation from our data-set.The parenthesis is given to represent EDU's.}
% \label{table:intraexampl}
% \end{table}

Rhetorical Structure Theory was originally developed to offer an explanation of the coherence of texts. \newcite{musi} and, more recently \newcite{hewett-etal-2019-utility}, showed that discourse relations from RST often correlate with argumentative relations. We thus derive features from RST trees and train a classifier using these features to predict an argumentative relation.  To extract features from a pair of argumentative components, we first concatenate the two components so that they form a single text input. We then use a state-of-the-art RST discourse parser \citep{rst}\footnote{We  use \newcite{wang2018edu} for segmentation of text into elementary discourse units.} % as they gain improved results using ELMO \newcite{elmo}}.
to create parse trees and take the predicted discourse relation at the root of the parse tree as a categorical feature in a classifier. %CH - add example?
There are 28 unique discourse relations predicted in the data, including \textit{Circumstance}, \textit{Purpose}, and \textit{Antithesis}.
We use a one-hot encoding of these relations as features and train an XGBoost Classifier \cite{xgb} to predict whether an argument relation exists. This classifier with discourse relations, as indicated in Figure \ref{modelpic}, is then ensembled with our predictions  from the BERT classifier by predicting a relation if either one of the classifiers predicts a relation.

\paragraph{Candidate Target Selection}
%KM2 This section is much better now.
For inter-turn relations, we take additional steps to reduce the number of invalid relation pairs.
Predicting an argumentative relation is made more difficult by the fact that we need to consider all possible relation pairs.
However, some argumentative components may contain linguistic properties that allow us to predict when they are targets even without the full relation pair.
Thus, if we can predict the targets with high recall, we are likely to increase precision as we can reduce the number of false positives.
Our candidate selection component, which identifies potential targets (as shown in Figure \ref{modelpic}), consists of two sub-components: an extractive summarizer and a source-target constraint.

\begin{figure}[h]
\centering
\includegraphics[scale=0.2, trim={0 1cm 0 0},]{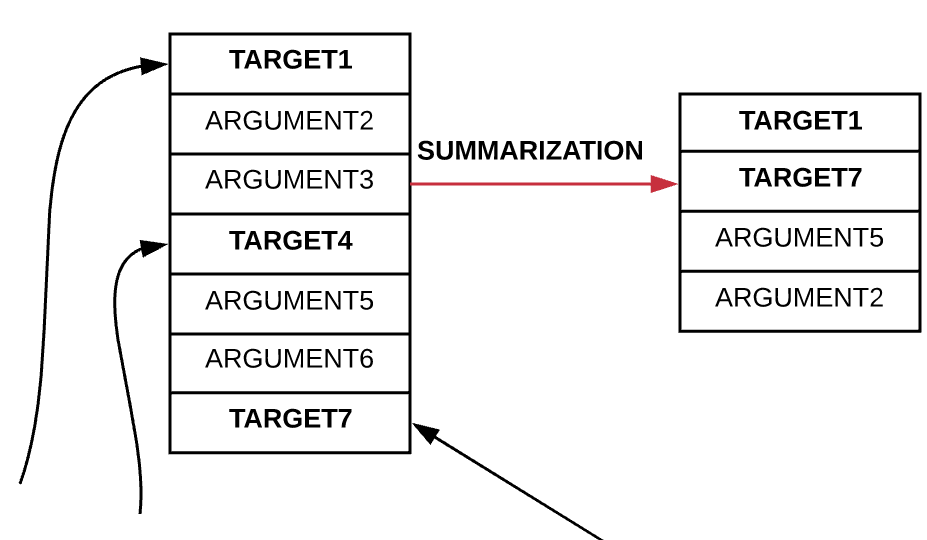}
\caption{\label{schematic-summ}Schematic of our Target Extraction Approach}
\label{target_extraction}
\end{figure}

First, we use the \textbf{QR} data to train a model to identify candidate targets using techniques from extractive summarization, with the idea that targets may be salient sentences or propositions. We treat the quoted sentences as gold labels, resulting in 19,413 pairs of document (post) and summary (quoted sentences).  For the example provided in Table \ref{summdata}, this would result in one sentence included in the summary. Thus, for a candidate argument pair A $\rightarrow$ B, where B is the quoted sentence in Table \ref{summdata}, if B is not extracted by the summarization model we predict that there is no relation between A and B. An example of target selection via extractive summarization is shown in Figure \ref{target_extraction}.

We use a state-of-the-art extractive summarization approach \citep{bertsum} for extracting the targets. The authors obtain sentence representations from BERT \citep{bert}, and build  several summarization specific layers stacked on top of the BERT outputs, to capture document-level features for extracting summaries. We refer the reader to \cite{bertsum} for further details.
We select the best summarization model on a held out subset using recall at the top $K$ sentences.
 
Second, in addition to summarization, we take advantage of a dataset-specific constraint: a target cannot also be a source unless it is related to the main claim. In other words, if B is a predicted target in A $\rightarrow$ B, we predict that there is no relation for B $\rightarrow$ A except when A is a main claim. In the CMV data, the main claim is always the title of the original Reddit post, so it is trivial to identify.

\begin{table} [h!]
\small
\centering
\begin{tabular}{|l|l|l|l|}
\hline
Method & C & P & NA \\ \hline
\newcite{pe} + EWE     & 56.0  & 65.9    & 69.6    \\ \hline
\newcite{japanese}      & 54.2  & 68.5    & 73.2    \\ \hline
\newcite{imho}      & 57.8  & 70.8    & 70.5    \\ \hline
BERT \newcite{bert}       & 62.0  & 72.2    & 71.3    \\ \hline
IMHO Context Fine-Tuned BERT     & \textbf{67.1}  & \textbf{72.5}    & \textbf{75.7}    \\ \hline
\end{tabular}
\caption{F-scores for 3-way Classification: Claim (C), Premise (P), Non-Argument (NA)}
\label{table:argfscore}
\end{table}

\begin{table*}[]
\center
\begin{tabular}{|l|l|l|l|l|l|l|}
\hline
\multirow{2}{*}{Method} & \multicolumn{2}{l|}{Precision} & \multicolumn{2}{l|}{Recall} & \multicolumn{2}{l|}{F-Score} \\ \cline{2-7} 
                        & Gold           & Pred          & Gold          & Pred        & Gold          & Pred         \\ \hline
All relations           & 5.0            & -             & 100.0         & -           & 9.0           & -            \\ \hline
\newcite{aaai18}                  & 7.0            & 5.9           & 82.0          & 80.0        & 13.0          & 11.0         \\ \hline
\newcite{aaai18} + RST Features              & 7.4            & 6.1           & 83.0          & 81.0        & 13.7          & 11.4         \\ \hline
RST Features                     & 6.3            & 5.7           & 79.5          & 77.0        & 11.8          & 10.6         \\ \hline
\newcite{japanese}            & 10.0           & -             & 48.8          & -           & 16.6          & -            \\ \hline
BERT  \newcite{bert}                  & 12.0           & 11.0          & 67.0          & 60.0        & 20.3          & 18.5         \\ \hline \hline
IMHO Context Fine-Tuned BERT                    & 14.3           & 13.2          & 69.0          & 65.0        & 23.7          & 21.8         \\ \hline
+ RST Ensemble                     & \textbf{16.7}          & \textbf{15.5}          & \textbf{73.0}          & \textbf{70.2}        & \textbf{27.2}          & \textbf{25.4}         \\ \hline
\end{tabular}
\caption{Results for Intra-turn Relation Prediction  with Gold and Predicted Premises}
\label{table:argrelationIPRfscore}
\end{table*}

\section{Experiments and Results} \label{results}
%KM - I think you should repeat the names of the datasets rather than just referring to the previous section. It is hard for reviewers and they forget. Make it easy on them.
We use the CMV data from Section \ref{labeled_data} for training and testing, setting aside 10\% of the data for test. %To make our experiments compatible with previous work, we train all baselines at the sentence level. 
Hyper-parameters are discussed in Appendix A. 

\subsection{Argumentative Component Classification}
\label{acc}

For baseline experiments on argumentative component classification we rely on the  handcrafted features used by \newcite{pe}: lexical (unigrams), structural (token statistics and position), indicator (\textit{I, me, my}),  syntactic (POS, modal verbs), discourse relation (PDTB), and word embedding features. \newcite{cmv} show that emotional appeal or pathos is strongly correlated with persuasion and appears in premises. This motivated us to augment the work of \newcite{pe} with emotion embeddings \citep{emoem} which capture emotion-enriched word representations and show improved performance over generic embeddings (denoted in the table as EWE).

We also compare our results to several neural models - a joint model using pointer networks \citep{japanese}, a model that leverages context fine-tuning \citep{imho}, and a BERT baseline \citep{bert} using only the pre-trained model without our additional fine-tuning step.

Table \ref{table:argfscore} shows that our best model gains statistically significant improvement over all the other models ($p  < 0.001$ with a Chi-squared test). To compare directly to the work of \newcite{imho}, we also test our model on the binary claim detection task and obtain a Claim F-Score of 70.0 with fine-tuned BERT, which is a 5-point improvement in F-score over pre-trained BERT and a 12-point improvement over \newcite{imho}. These results show that fine-tuning on the appropriate context is key.

\begin{table*}[]
\center
\begin{tabular}{|l|l|l|l|l|l|l|}
\hline
\multirow{2}{*}{Method} & \multicolumn{2}{l|}{Precision} & \multicolumn{2}{l|}{Recall} & \multicolumn{2}{l|}{F-Score} \\ \cline{2-7} 
                        & Gold           & Pred          & Gold          & Pred        & Gold          & Pred         \\ \hline
All relations           & 5.0            & -             & 100.0         & -           & 9.0           & -            \\ \hline
\newcite{aaai18}                  & 5.9            & 4.8           & 82.0          & 80.0        & 11.0          & 9.0          \\ \hline
\newcite{aaai18} + RST Features              & 6.4            & 4.9           & 83.0          & 80.0        & 11.8          & 9.3          \\ \hline
RST Features                     & 5.1            & 3.8           & 80.0          & 77.0        & 9.6           & 7.2          \\ \hline
\newcite{japanese}            & 7.6            & -             & 40.0          & -           & 12.7          & -            \\ \hline
BERT \newcite{bert}                    & 8.8            & 7.9           & 76.0          & 70.0        & 15.8          & 14.1         \\ \hline \hline
QR Context Fine-Tuned BERT                      & 11.0           & 10.0          & 75.3          & 72.5        & 19.1          & 17.6         \\ \hline
+ RST Features                     & 11.0           & 12.2          & 79.0          & 75.5        & 21.2          & 19.1         \\ \hline
+ Extractive Summarizer                    & 16.0           & 14.5          & \textbf{79.4}          & \textbf{75.6}        & 26.8          & 24.3         \\ \hline
+ Source $\neq$ Target  Constraint              & \textbf{18.9}           & \textbf{17.5}         & 79.0          & 74.0        & \textbf{30.5}          & \textbf{28.3}         \\ \hline
\end{tabular}
\caption{Results for Inter-Turn Relation Prediction with Gold and Predicted Claims}
\label{table:argrelationIPIfscore}
\end{table*}

\subsection{Relation Prediction}
For baseline experiments on relation prediction, we consider prior work in macro-level argument mining.
\newcite{aaai18} predict argumentative relations between entire political speeches from different speakers, which is similar to our dialogues. 
We re-implement their model using their features (lexical overlap, negation, argument entailment, and argument sentiment, among others).
As with component classification, we also compare to neural models for relation prediction - the joint pointer network architecture \citep{japanese} and the pre-trained BERT \citep{bert} baseline. 

As the majority of component pairs contain no relation, we could obtain high accuracy by predicting that all pairs have no relation.  Instead, we want to measure our performance on relations, so we also include an ``all-relation'' baseline, where we always predict that there is a relation between two components, to indicate the difficulty of modeling such an imbalanced data set. In the test data, for intra-turn relations there are 2264 relation pairs, of which only 174 have a relation, and for inter-turn relations there are 120 relation pairs, compared to 2381 pairs with no relation. 

As described in Section \ref{labeled_data}, for intra-turn relations, the source is constrained to be a premise whereas for intra-turn, it is constrained to be a claim.  We thus provide experiments using both gold claims/premises and predicted ones.

\paragraph{Intra-turn Relations}
We report the results of our binary classification task in Table \ref{table:argrelationIPRfscore} in terms of Precision, Recall and F-score for the ``true'' class, i.e., when a relation is present. We report results given both gold premises and predicted premises (using our best model from \ref{acc}). Our best results are obtained from ensembling the RST classifier with BERT fine-tuned on \textbf{IMHO+context}, for statistically significant ($p<0.001$) improvement over all other models. 
We obtain comparable performance to previous work on relation prediction in other argumentative datasets \citep{factor-graph, japanese}.

\paragraph{Inter-turn Relations}

As with intra-turn relations, we report F-score on the ``true'' class in Table \ref{table:argrelationIPIfscore} for both gold and predicted claims.
Our best results are obtained by fine-tuning the BERT model on the appropriate context (in this case the \textbf{QR} data) and ensembling the predictions with the RST classifier. We again obtain statistically significant ($p<0.001$) improvement over all baselines.

Our methods for candidate target selection obtain further improvement. Using our extractive summarizer, we found that we obtained the best target recall of 62.7 at $K=5$ (the number of targets to select). This component improves performance by approximately 5 points in F-score by reducing the search space of relation pairs.
By further constraining targets to only be a source when targeting a main claim, we obtain another 4 point gain.

\begin{figure}[h]
\centering
\includegraphics[scale=0.35, trim={0 1cm 0 1cm},]{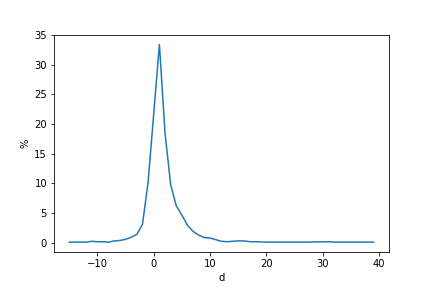}
\caption{\label{dist}Distances $d$ between Intra-Turn Relations}
\label{distances}
\end{figure}

\paragraph{Window Clipping}
We also conduct experiments showing the performance for intra-turn relation prediction when constraining the relations to be within a certain distance in terms of the number of sentences apart.  Often, in persuasive essays or within a single post the authors use premises to back/justify claims they immediately made. As shown in Figure \ref{dist}, this behavior is also reflected in our dataset where the distance between the two arguments in the majority of the relations is +1. % (the argumentative component immediately following). 

We thus limit the model's prediction of a relation to be within a certain window and predict ``no relation'' for any pairs outside of that window.
Table \ref{table:windowclip} shows that this window clipping on top of our best model improves F-score by only limiting the context where we make predictions. As our models are largely trained on discourse context and the next sentence usually has a discourse relation, we obtain improved performance as we narrow the window size. While we see a drop in recall  the precision improves compared to our previous results in Table \ref{table:argrelationIPRfscore}. It is also important to note that window clipping is only beneficial once we have a high recall, low precision scenario because when we predict everything at a distance of +1 as a relation we obtain low F-scores.

\begin{table*}[]
\center
\begin{tabular}{|l|l|l|l|l|l|l|l|}
\hline
\multirow{2}{*}{Method}     & \multirow{2}{*}{Window} & \multicolumn{2}{l|}{Precision} & \multicolumn{2}{l|}{Recall} & \multicolumn{2}{l|}{F-Score} \\ \cline{3-8} 
                            &                         & Gold           & Pred          & Gold         & Pred         & Gold          & Pred         \\ \hline
All relations               & 0 TO +1                 & 5.0            & 4.0           & 31.0         & 25.0         & 8.7           & 6.9          \\ \hline
\multirow{5}{*}{Best Model} & 0 TO +5                 & 19.5           & 17.1          & \textbf{70.0}         & \textbf{67.0}         & 30.5          & 27.2         \\ \cline{2-8} 
                            & 0 TO +4                 & 21.4           & 19.5          & 67.0         & 65.0         & 32.2          & 30.0         \\ \cline{2-8} 
                            & 0 TO +3                 & 25.2           & 23.3          & 61.1         & 58.0         & 35.6          & 33.2         \\ \cline{2-8} 
                            & 0 TO +2                 & 32.5           & 29.8          & 50.0         & 48.0         & 39.3          & 36.8         \\ \cline{2-8} 
                            & 0 TO +1                 & \textbf{41.5}           & \textbf{39.1}          & 47.0         & 42.0         & \textbf{44.1}          & \textbf{40.3}         \\ \hline
\end{tabular}
 \caption{Intra-Turn Relation Prediction with Varying Window Settings}
% %. All relations is a baseline comparison which predicts everything at a distance of +1 as a relation.
\label{table:windowclip}
\end{table*}

% \begin{table} [h!]
% \small
% \centering
% \begin{tabular}{|l|l|l|c|}
% \hline
% \multirow{2}{*}{Method}                                                & \multirow{2}{*}{Window} & \multicolumn{2}{l|}{F-score} \\ \cline{3-4} 
%                                                                       &                         & Gold          & Pred         \\ \hline
%                                                                       \begin{tabular}[c]{@{}l@{}}All relations\end{tabular}            & 0 TO +1                 & 8.7           & 6.5          \\ \hline
% \multirow{5}{*}{\begin{tabular}[c]{@{}l@{}}Best \\ Model\end{tabular}} & 0 TO +5                 & 30.5          & 27.2         \\ \cline{2-4} 
%                                                                       & 0 TO +4                 & 32.3          & 30.0         \\ \cline{2-4} 
%                                                                       & 0 TO +3                 & 35.6          & 33.2         \\ \cline{2-4} 
%                                                                       & 0 TO +2                 & 39.3          & 36.8         \\ \cline{2-4} 
%                                                                       & 0 TO +1                 & \textbf{44.1}          & \textbf{40.3}         \\ \hline

% \end{tabular}
% \caption{Intra-Turn Relation Prediction with Varying Window Settings}
% %. All relations is a baseline comparison which predicts everything at a distance of +1 as a relation.
% \label{table:windowclip}
% \end{table}

% %\vspace{-3ex}
\section{Qualitative Analysis}
\paragraph{Role of Context}

We retrieve examples from the \textbf{IMHO+context} and \textbf{QR} data using TF-IDF similarity to pairs of argument components from our data that were predicted incorrectly by pre-trained BERT but correctly by the respective fine-tuned model.
The first two rows in Table \ref{table:erroran} show a relation between a claim and premise in the IMHO+Context and the CMV data respectively while the last four rows show a relation between a claim and premise in the QR data and the CMV data. The model learns discriminative discourse relations from the \textbf{IMHO+context} data and correctly identifies this pair. The last four rows show rebuttal from the QR and the CMV data respectively, where the model learns discriminative dialogic phrases (highlighted in bold).

\begin{table} [h!]
\small
\centering
\begin{tabular}{|p{0.9cm}|p{6cm}|}
\hline
Context & Pair \\
\hline
IMHO & IMHO, you should not \textbf{quantify} it as good or bad.
\textbf{Tons of people} have monogamous relationships without issue. \\ \hline
CMV                                                        & [how would you even \textbf{quantify} that] [there are \textbf{many people} who want close relationships without romance]          \\ 
\hline
\hline

QR & [It might be that egalitarians,anti-feminists, MRAs \& redpillers, groups that I associate with opposing feminism - might be in fact very \textbf{distinct} \& \textbf{different} groups,  but I don't know that] [\textbf{I do see} all four of these as \textbf{distinct} groups]. \\ 
\hline
CMV &  [I may have a different stance on seeing no \textbf{difference} between companion animals and farm animals.] [\textbf{I do see distinction} between a pet and livestock]   \\ 
\hline
\hline
QR & [Of course you \textbf{intend to kill} the person if you draw your \textbf{weapon}, if you can reasonably assume that they have a \textbf{weapon}] [\textbf{I don't think} some of them would \textbf{start killing}]. \\  \hline
 CMV &  [So i thought, why would a police officer even use \textbf{firearms} if he/she doesn't \textbf{intend to kill}?] [\textbf{I don’t think}, police are allowed to \textbf{start killing} someone with their gun if they don't intend to . ]   \\
 \hline
\end{tabular}
\caption{CMV and Context Examples}
\label{table:erroran}
\end{table}

%%\vspace{-2ex}
\paragraph{Role of Discourse}

\begin{table} [h!]
\small
\centering
\begin{tabular}{|l|l|l|}
\hline
Discourse                                                         & Argument1                                                                                                                         & Argument2                                                                                                                                                  \\ \hline
\begin{tabular}[c]{@{}l@{}}Evaluation\end{tabular} & \begin{tabular}[c]{@{}l@{}}The only way your\\ life lacks meaning\\ is if you give it none\\ to begin with\end{tabular}       & \begin{tabular}[c]{@{}l@{}}Life is ultimately\\ meaningless and \\ pointless.\end{tabular}                                                            \\ \hline
\begin{tabular}[c]{@{}l@{}}Antithesis\end{tabular}  & \begin{tabular}[c]{@{}l@{}}Joseph was just a\\ regular Jew without\\ the same kind of \\holiness as the \\other two\end{tabular} & \begin{tabular}[c]{@{}l@{}}Aren't Mary and\\ Joseph, two holy\\ people especially \\perfect virgin Mary,\\both Jews? Wasn't\\ Jesus a Jew?\end{tabular} \\ \hline
\end{tabular}
\caption{Predicted Discourse Relations in CMV}
\label{table:discan}
\end{table}

We also provide examples that are predicted incorrectly by BERT but correctly by our classifier trained with RST features. 
For the first example in Table \ref{table:discan} the RST parser predicts an \textit{Evaluation} relation, which is an indicator of an argumentative relation according to our model. 
For the second example the RST parser predicts \textit{Antithesis}, which is correlated with attack relations \citep{musi}, and is predicted correctly by our model.

\section{Conclusion}
We show how fine-tuning on data-sets similar to the task of interest is often beneficial. As our data set is small we demonstrated how to use transfer learning by leveraging discourse and dialogue context. We show how the structure of the fine-tuning corpus is essential for improved performance on pre-trained language models. We also showed that predictions that take advantage of RST discourse cues are complementary to BERT predictions. Finally, we demonstrated methods to reduce the search space and improve precision.

In the future, we plan to experiment further with language model fine-tuning on other sources of data. We also plan to investigate additional RST features.  As the RST parser is not perfect, we want to incorporate other features from these trees that allow us to better recover from errors. 
%SM-final. This sentence below does not make much sense to me. I commented it out
%Finally, we plan to investigate other tasks such as influence detection or argument generation as discourse structure is a key factor in persuasion.\linebreak

\section{Acknowledgements}
The authors thank Fei-Tzin Lee and Alexander Fabbri for their helpful comments on the initial draft of this paper and the anonymous reviewers for helpful comments.

\bibliography{emnlp-ijcnlp-2019}

\begin{thebibliography}{35}
\expandafter\ifx\csname natexlab\endcsname\relax\def\natexlab#1{#1}\fi

\bibitem[{Abbott et~al.(2011)Abbott, Walker, Anand, Fox~Tree, Bowmani, and
  King}]{abbott2011can}
Rob Abbott, Marilyn Walker, Pranav Anand, Jean~E Fox~Tree, Robeson Bowmani, and
  Joseph King. 2011.
\newblock How can you say such things?!?: Recognizing disagreement in informal
  political argument.
\newblock In \emph{Proceedings of the Workshop on LSM}, pages 2--11.

\bibitem[{Agrawal et~al.(2018)Agrawal, An, Chen, and Manos}]{emoem}
Ameeta Agrawal, Aijun An, Papagelis Chen, and Manos. 2018.
\newblock Learning emotion-enriched word representations.
\newblock In \emph{Proceedings of the 27th International Conference on
  Computational Linguistics}.

\bibitem[{Bentahar et~al.(2010)Bentahar, Moulin, and
  B{\'e}langer}]{bentaharetal2010}
Jamal Bentahar, Bernard Moulin, and Micheline B{\'e}langer. 2010.
\newblock A taxonomy of argumentation models used for knowledge representation.
\newblock \emph{Artif. Intell. Rev.}, 33(3):211--259.

\bibitem[{Boltu{\v{z}}i{\'c} and {\v{S}}najder(2014)}]{stancebackup}
Filip Boltu{\v{z}}i{\'c} and Jan {\v{S}}najder. 2014.
\newblock Back up your stance: Recognizing arguments in online discussions.
\newblock In \emph{Proceedings of the First Workshop on Argumentation Mining},
  pages 49--58.

\bibitem[{Budzynska et~al.(2014)Budzynska, Janier, Kang, Reed, Saint-Dizier,
  Stede, and Yaskorska}]{budzynska2014towards}
Katarzyna Budzynska, Mathilde Janier, Juyeon Kang, Chris Reed, Patrick
  Saint-Dizier, Manfred Stede, and Olena Yaskorska. 2014.
\newblock Towards argument mining from dialogue.

\bibitem[{Chakrabarty et~al.(2019)Chakrabarty, Hidey, and McKeown}]{imho}
Tuhin Chakrabarty, Christopher Hidey, and Kathleen McKeown. 2019.
\newblock Imho fine-tuning improves claim detection.
\newblock \emph{arXiv preprint arXiv:1905.07000}.

\bibitem[{Chen and Guestrin(2016)}]{xgb}
Tianqi Chen and Carlos Guestrin. 2016.
\newblock \href {https://doi.org/10.1145/2939672.2939785} {{XGBoost}: A
  scalable tree boosting system}.
\newblock In \emph{Proceedings of the 22nd ACM SIGKDD International Conference
  on Knowledge Discovery and Data Mining}, KDD '16, pages 785--794, New York,
  NY, USA. ACM.

\bibitem[{Devlin et~al.(2019)Devlin, Chang, Kenton, and Toutanova}]{bert}
Jacob Devlin, Ming-Wei Chang, Lee Kenton, and Kristina Toutanova. 2019.
\newblock Bert: Pre-training of deep bidirectional transformers for language
  understanding.
\newblock In \emph{Proceedings of the 17th Annual Meeting of the North American
  Association for Computational Linguistics}.

\bibitem[{Eger et~al.(2017)Eger, Daxenberger, and Gurevych}]{end2end}
Steffen Eger, Johannes Daxenberger, and Iryna Gurevych. 2017.
\newblock Neural end-to-end learning for computational argumentation mining.
\newblock In \emph{In Proceedings of the 55th Annual Meeting of the Association
  for Computational Linguistics.}, pages 11–--22.

\bibitem[{Feng and Hirst(2011)}]{feng2011}
Vanessa~Wei Feng and Graeme Hirst. 2011.
\newblock Classifying arguments by scheme.
\newblock In \emph{ACL}, pages 987--996.

\bibitem[{Ghosh et~al.(2014)Ghosh, Muresan, Nina, Aakhus, and Mitsui}]{target}
Debanjan Ghosh, Smaranda Muresan, Wacholder Nina, Mark Aakhus, and Matthew.
  Mitsui. 2014.
\newblock Analyzing argumentative discourse units in online interactions.
\newblock In \emph{In Proceedings of the First Workshop on Argument Mining,
  hosted by the 52nd Annual Meeting of the Association for Computational
  Linguistics, ArgMining@ACL}, pages 39–--48.

\bibitem[{Habernal and Gurevych(2017)}]{habernalgurevych2017}
Ivan Habernal and Iryna Gurevych. 2017.
\newblock Argumentation mining in user-generated web discourse.
\newblock \emph{Computational Linguistics}, 43(1):125--179.

\bibitem[{Hewett et~al.(2019)Hewett, Prakash~Rane, Harlacher, and
  Stede}]{hewett-etal-2019-utility}
Freya Hewett, Roshan Prakash~Rane, Nina Harlacher, and Manfred Stede. 2019.
\newblock \href {https://www.aclweb.org/anthology/W19-4512} {The utility of
  discourse parsing features for predicting argumentation structure}.
\newblock In \emph{Proceedings of the 6th Workshop on Argument Mining}, pages
  98--103, Florence, Italy. Association for Computational Linguistics.

\bibitem[{Hidey et~al.(2017)Hidey, Musi, Hwang, Muresan, and McKeown}]{cmv}
Christopher Hidey, Elena Musi, Alyssa Hwang, Smaranda Muresan, and Kathleen
  McKeown. 2017.
\newblock Analyzing the semantic types of claims and premises in an online
  persuasive forum.
\newblock In \emph{In Proceedings of the 4th Workshop on Argument Mining.
  EMNLP}, pages 11–--21.

\bibitem[{Ji and Eisenstein(2014)}]{rst}
Yangfeng Ji and Jacob Eisenstein. 2014.
\newblock Representation learning for text-level discourse parsing.
\newblock In \emph{Proceedings of the 52nd Annual Meeting of the Association
  for Computational Linguistics,}, pages 13--24.

\bibitem[{Liu(2019)}]{bertsum}
Yang Liu. 2019.
\newblock Fine-tune bert for extractive summarization.
\newblock In \emph{https://arxiv.org/abs/1903.10318}.

\bibitem[{Mann and Thompson(1988)}]{mannRST}
William~C Mann and Sandra~A Thompson. 1988.
\newblock Rhetorical structure theory: Toward a functional theory of text
  organization.
\newblock \emph{Text-Interdisciplinary Journal for the Study of Discourse},
  8(3):243--281.

\bibitem[{Menini et~al.(2018)Menini, Cabrio, Tonelli, and Serena}]{aaai18}
Stefano Menini, Elena Cabrio, Sara Tonelli, and Villata Serena. 2018.
\newblock Never retreat, never retract: Argumentation analysis for political
  speeches.
\newblock In \emph{Association for the Advancement of Artificial Intelligence}.

\bibitem[{Morio and Fujita(2018)}]{japanese}
Gaku Morio and Katsuhide Fujita. 2018.
\newblock End-to-end argument mining for discussion threads based on parallel
  constrained pointer architecture.
\newblock In \emph{In Proceedings of the 5th Workshop on Argument Mining.
  EMNLP}, pages 11--21.

\bibitem[{Murakami and Raymond(2010)}]{murakami}
Akiko Murakami and Rudy Raymond. 2010.
\newblock Support or oppose?: Classifying positions in online debates from
  reply activities and opinion expressions.
\newblock In \emph{In Proceedings of the 23rd International Conference on
  Computational Linguistics, ArgMining@ACL}, pages 869–--875.

\bibitem[{Musi et~al.(2018)Musi, Alhindi, Manfred, Kriese, Muresan, and
  Rocci}]{musi}
Elena Musi, Tariq Alhindi, Stede Manfred, Leonard Kriese, Smaranda Muresan, and
  Andrea Rocci. 2018.
\newblock A multi-layer annotated corpus of argumentative text: From argument
  schemes to discourse relations.
\newblock In \emph{Proceedings of Language Resources and Evaluation Conference
  (LREC 2018)}.

\bibitem[{Niculae et~al.(2017)Niculae, Park, and Cardie}]{factor-graph}
Vlad Niculae, Joonsuk Park, and Claire Cardie. 2017.
\newblock \href {https://doi.org/10.18653/v1/P17-1091} {Argument mining with
  structured {SVM}s and {RNN}s}.
\newblock In \emph{Proceedings of the 55th Annual Meeting of the Association
  for Computational Linguistics (Volume 1: Long Papers)}, pages 985--995,
  Vancouver, Canada. Association for Computational Linguistics.

\bibitem[{Peldszus and Stede(2015)}]{peldszusstede2015}
Andreas Peldszus and Manfred Stede. 2015.
\newblock Joint prediction in mst-style discourse parsing for argumentation
  mining.
\newblock In \emph{EMNLP}, pages 938--948.

\bibitem[{Persing and Ng(2016)}]{essay}
Isaac Persing and Vincent Ng. 2016.
\newblock End-to-end argumentation mining in student essays.
\newblock In \emph{In Proceedings of the 15th Conference of the North American
  Chapter of the Association for Computational Linguistics: Human Language
  Technologies.}, pages 1384--1394.

\bibitem[{Potash et~al.(2017)Potash, Romanov, and Rumshisky}]{potash}
Peter Potash, Alexey Romanov, and Anna Rumshisky. 2017.
\newblock Here’s my point: Joint pointer architecture for argument mining.
\newblock In \emph{Proceedings of the 2017 Conference on Empirical Methods in
  Natural Language Processing}, pages 1364--1373.

\bibitem[{Rosenthal and McKeown(2015)}]{rosenthal2015couldn}
Sara Rosenthal and Kathleen McKeown. 2015.
\newblock I couldn't agree more: The role of conversational structure in
  agreement and disagreement detection in online discussions.
\newblock In \emph{16th Annual Meeting of the SIGDIAL}.

\bibitem[{Somasundaran et~al.(2007)Somasundaran, Ruppenhofer, and
  Wiebe}]{somasundaran2007detecting}
Swapna Somasundaran, Josef Ruppenhofer, and Janyce Wiebe. 2007.
\newblock Detecting arguing and sentiment in meetings.
\newblock In \emph{Proceedings of the SIGdial Workshop on Discourse and
  Dialogue}, volume~6.

\bibitem[{Sridhar et~al.(2015)Sridhar, Foulds, Huang, Getoor, and
  Walker}]{sridhar2015joint}
Dhanya Sridhar, James~R Foulds, Bert Huang, Lise Getoor, and Marilyn~A Walker.
  2015.
\newblock Joint models of disagreement and stance in online debate.
\newblock In \emph{ACL}, pages 116--125.

\bibitem[{Stab and Gurevych(2014)}]{stab2014identifying}
Christian Stab and Iryna Gurevych. 2014.
\newblock Identifying argumentative discourse structures in persuasive essays.
\newblock In \emph{EMNLP}.

\bibitem[{Stab and Gurevych(2017)}]{pe}
Christian Stab and Iryna Gurevych. 2017.
\newblock Parsing argumentation structures in persuasive essays.
\newblock In \emph{Computational Linguistics, pages in press, preprint
  available at arXiv:1604.07370}.

\bibitem[{Swanson et~al.(2015)Swanson, Ecker, and Walker}]{swanson2015argument}
Reid Swanson, Brian Ecker, and Marilyn Walker. 2015.
\newblock Argument mining: Extracting arguments from online dialogue.
\newblock In \emph{Proceedings of the 16th Annual Meeting of the SIGDIAL},
  pages 217--226.

\bibitem[{Vinyals et~al.(2015)Vinyals, Fortunato, and Jaitly}]{pointer2015}
Oriol Vinyals, Meire Fortunato, and Navdeep Jaitly. 2015.
\newblock Pointer networks.
\newblock In \emph{In C. Cortes, N. D. Lawrence, D. D. Lee, M. Sugiyama, and R.
  Garnett, editors, Advances in Neural Information Processing Systems 28},
  pages 2692–--2700.

\bibitem[{Visser et~al.(2019)Visser, Konat, Duthie, Koszowy, Budzynska, and
  Reed}]{Visser2019}
Jacky Visser, Barbara Konat, Rory Duthie, Marcin Koszowy, Katarzyna Budzynska,
  and Chris Reed. 2019.
\newblock \href {https://doi.org/10.1007/s10579-019-09446-8} {Argumentation in
  the 2016 us presidential elections: annotated corpora of television debates
  and social media reaction}.
\newblock \emph{Language Resources and Evaluation}.

\bibitem[{Walker et~al.(2012)Walker, Tree, Anand, Abbott, and
  King}]{walker2012corpus}
Marilyn~A Walker, Jean E~Fox Tree, Pranav Anand, Rob Abbott, and Joseph King.
  2012.
\newblock A corpus for research on deliberation and debate.
\newblock In \emph{LREC}, pages 812--817.

\bibitem[{Wang et~al.(2018)Wang, Li, and Yang}]{wang2018edu}
Yizhong Wang, Sujian Li, and Jingfeng Yang. 2018.
\newblock Toward fast and accurate neural discourse segmentation.
\newblock In \emph{Proceedings of the 2018 Conference on Empirical Methods in
  Natural Language Processing}, pages 962--967.

\end{thebibliography}
\bibliographystyle{acl_natbib}
\end{document}